%% file: document.tex
\documentclass[11pt]{article}

\usepackage[margin=1in]{geometry}
\usepackage{times}
\usepackage{amsmath, amssymb}
\usepackage{graphicx}
\usepackage{booktabs}
\usepackage{algorithm}
\usepackage{algorithmic}
\usepackage{hyperref}
\usepackage{natbib}
\usepackage{xcolor,pifont}
\usepackage{wrapfig}
\usepackage{multirow}
\usepackage{tikz}

\usepackage{iclr2027_conference}

\input{math_commands.tex}

\usepackage{url}

\newcommand*\colourcheck[1]{%
  \expandafter\newcommand\csname #1check\endcsname{\textcolor{#1}{\ding{52}}}%
}
\colourcheck{blue}
\colourcheck{green}
\colourcheck{red}

\newcommand*\colourcross[1]{%
  \expandafter\newcommand\csname #1cross\endcsname{\textcolor{#1}{\ding{56}}}%
}
\definecolor{citationgreen}{RGB}{0, 255, 0}
\hypersetup{
    colorlinks=true,
    citecolor=citationgreen,
    linkcolor=black,
    urlcolor=blue
}
\colourcross{red}

\title{
RepoMAS: Solving Progressively Specified Tasks with Issue-Driven Multi-Agent Systems
}

\author{
Yuchen Song\textsuperscript{1}\thanks{Equal contribution.},
Andong Chen\textsuperscript{1}\footnotemark[1],
Wenxin Zhu\textsuperscript{1},
\textbf{Muyun Yang}\textsuperscript{1},
\textbf{Tiejun Zhao}\textsuperscript{1}\thanks{Corresponding author.}
\\[2pt]
\textsuperscript{1}Harbin Institute of Technology, Harbin, China
\\
\texttt{songyuchn@126.com, ands691119@gmail.com, wenxinzhu@stu.hit.edu.cn}
\\
\texttt{\{yangmuyun, tjzhao\}@hit.edu.cn}
}
\iclrfinalcopy
\begin{document}

\maketitle

\renewcommand{\thefootnote}{\arabic{footnote}}
\setcounter{footnote}{0}

\begin{abstract}
LLM-based multi-agent systems (MASs) have shown strong potential for solving complex tasks, but most assume that task requirements are sufficiently specified before execution. In practice, user requests are often incomplete, and additional requirements may only become clear during reasoning, tool use, or execution. We refer to such problems as \textit{progressively specified tasks}.

To systematically study this setting, we introduce \textbf{ProgSpec}, a benchmark that evaluates final outputs against requirements explicitly stated in the initial request and additional requirements supported by the available task evidence. We further propose \textbf{RepoMAS}, an issue-driven multi-agent framework inspired by open-source project management. RepoMAS records newly discovered requirements, conflicts, and failures as structured Issues and uses them to revise the task specification and execution structure during problem solving.

Across ProgSpec and five existing benchmarks, RepoMAS achieves the best performance. Further analyses show that its issue-driven revision and repository maintenance mechanisms consistently contribute to performance. These results highlight the importance of allowing MASs to revise not only how a task is solved, but also revise their explicit representation of task requirements during execution.\footnote{\color{blue}{Our code and data will be available once the paper is accepted.}}
\end{abstract}

\setlength{\intextsep}{5pt}
\section{Introduction}

\begin{figure}[!h]
    \centering
    \includegraphics[width=\linewidth]{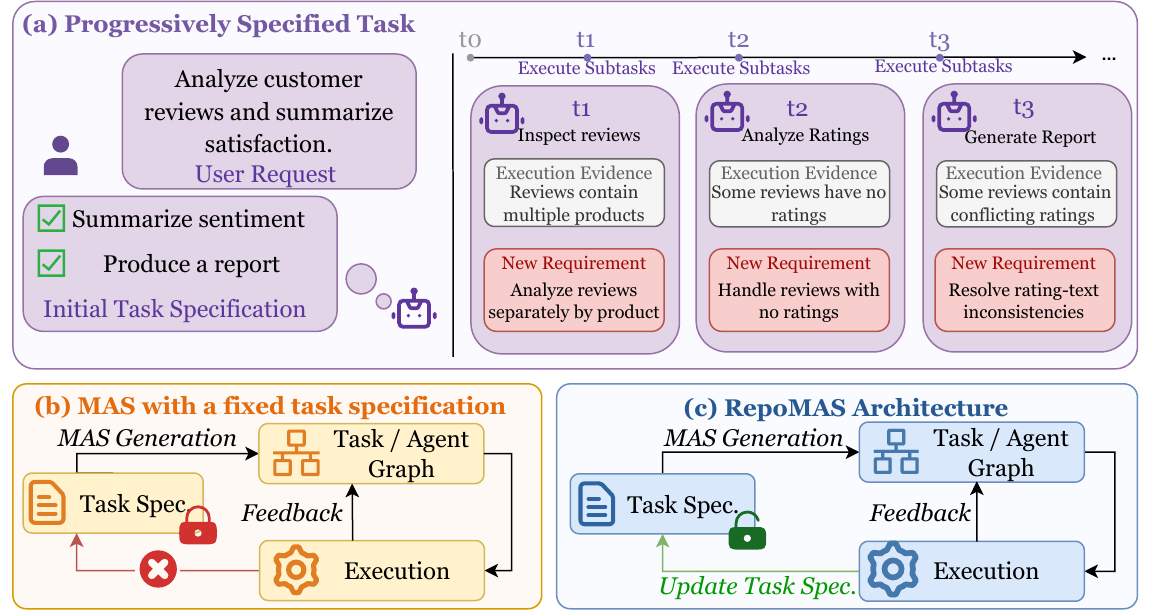}
\caption{\textbf{Overview of progressively specified tasks and RepoMAS.}
Task requirements may emerge gradually from execution evidence rather than being fully specified in the initial request. RepoMAS records these newly discovered requirements and revises the task specification and execution structure accordingly.}
    \label{intro_figure}
    \vspace{-0.1cm}
\end{figure}

LLM-based multi-agent systems (MASs) have become a promising paradigm for solving complex tasks. Instead of relying on a single model to complete an entire task, MASs decompose complex problems into subtasks and assign them to specialized agents. By organizing agents into specialized roles and enabling them to collaborate, these systems can handle complex tasks more flexibly and effectively~\citep{guolarge, hong2024metagpt, chen2024agentverse}.

While existing MASs have made substantial progress, most MASs implicitly assume that the task is sufficiently specified before execution begins, as shown in Figure~\ref{intro_figure}(b). Users usually describe an overall goal in a few sentences, but they may not be able to specify every requirement in advance. Some specifications remain latent until a tentative solution makes their absence visible, while others become explicit only after the system retrieves evidence, invokes tools, or interacts with the environment. As these issues are discovered, both the solution and the system's understanding of the task must be revised accordingly. As illustrated in Figure~\ref{intro_figure}(a), some task requirements become visible only during execution. We refer to such tasks as \emph{progressively specified tasks}.

To systematically study this setting, we introduce \textbf{ProgSpec}, a benchmark for progressively specified tasks. It evaluates whether a system can identify and handle specifications that are missing from the initial request and discovered during problem solving. To address progressively specified tasks, we propose \textbf{RepoMAS}, an issue-driven multi-agent framework that allows the task specification to be refined during problem solving. Its design is inspired by the workflow of open-source project management, where newly discovered problems are recorded as Issues, addressed through targeted revisions, and incorporated only after validation. 

Given an initial user request, RepoMAS first constructs a Task Draft that reflects its current understanding of the task. As execution proceeds, retrieved evidence, tool use, and intermediate results may reveal previously unspecified requirements. RepoMAS records these findings as structured Issues, which capture what has been discovered and what needs to be revised. These Issues first revise the Task Draft and trigger corresponding updates to the task structure and agent organization. After revising the repository state, RepoMAS locally re-executes only the affected tasks. This iterative process allows RepoMAS to start from an incomplete request and progressively refine its understanding. A simplified overview of RepoMAS is shown in Figure~\ref{intro_figure}(c).

In summary, this work makes three main contributions:
\begin{enumerate}
    \item We formulate progressively specified tasks and introduce ProgSpec, a benchmark for evaluating whether systems satisfy requirements that are not explicitly stated in the initial request.
    \item We propose RepoMAS, an issue-driven multi-agent framework that turns execution feedback into structured Issues, which guide revisions to the current task specification.
    \item With GPT-4o-mini as the backbone, RepoMAS outperforms the evaluated baselines on ProgSpec and five established benchmarks. Experiments with additional backbones further demonstrate its applicability across different model families and tasks.
\end{enumerate}

\section{Related Work}

\subsection{LLM-based Multi-Agent Systems}

LLM-based multi-agent systems solve complex tasks through role specialization, communication, and coordinated workflows~\citep{guolarge, he2025llm, tran2025multi}. Representative systems,  such as CAMEL~\citep{li2023camel}, DyLAN~\citep{liu2023dynamic}, and AgentVerse~\citep{chen2024agentverse},  study how multiple agents collaborate through natural-language interaction.

Later work further treats multi-agent organization as a structure that can be explicitly designed and optimized. MacNet~\citep{qian2025scaling} studies large-scale collaboration through communication graphs, while AgentPrune~\citep{zhang2025cut} removes redundant connections to reduce communication cost. Together, these studies examine how communication structure affects collaboration and its computational cost.

\subsection{Dynamic Task Planning and Agent Collaboration}

Recent works move beyond fixed task plans toward task-dependent and execution-time adaptation. AutoAgents~\citep{chenautoagents} jointly constructs task-specific agents and execution plans, while MaAS~\citep{zhang2025multi} selects multi-agent architectures according to task complexity. Other approaches~\citep{yu2025dyntaskmas, wang2026dyflow, xu2026meta, li2026coupled} revise task decomposition, dependencies, or execution structures based on intermediate results and failures.

Agent organization can also be adapted during execution. Evolving Orchestration~\citep{dang2026multi} and AgentConductor~\citep{wang2026agentconductor} adjust agent activation or interaction structures, while other methods~\citep{ruan2026aorchestra, wang2026metagen, xu2026tacomas} further revise agent roles, capabilities, or communication patterns.

RepoMAS maintains explicit requirement--task links. Newly identified constraints can trigger coordinated revisions to the task specification and execution structures. These revisions are managed through structured Issues and validated patches.

\subsection{Benchmarks for Agents and Incomplete Specifications}

\begin{table}[!ht]
    \centering
    \vspace{-0.3cm}
        \caption{\textbf{Comparison between ProgSpec and related agent benchmarks.}
    Autonomous requirement recovery denotes recovery from reasoning or execution evidence rather than requirements supplied through user clarification.}

   \resizebox{\linewidth}{!}{

    \begin{tabular}{lccc}
    \toprule
    \textbf{Benchmark} 
    & \shortstack{\textbf{Incomplete}\\\textbf{Specifications}}
    & \shortstack{\textbf{Autonomous}\\\textbf{Requirement Recovery}}
    & \textbf{Multi-Domain} \\
    \midrule
    AgentBench~\citep{liu2024agentbench}
    & \redcross & \redcross & \greencheck \\

    GAIA~\citep{mialon2024gaia}
    & \redcross & \redcross & \greencheck \\

    MultiAgentBench~\citep{zhu2025multiagentbench}
    & \redcross & \redcross & \greencheck \\

    SWE-Together~\citep{wu2026swe}
    & \greencheck & \redcross & \redcross \\

    ICAE-Bench~\citep{peng2026icae}
    & \greencheck & \redcross & \redcross \\
    SpecBench~\citep{hamblin2026specbench}
    & \greencheck & \greencheck & \redcross \\

    \textbf{ProgSpec (Ours)}
    & \greencheck & \greencheck & \greencheck \\
    \bottomrule
    \end{tabular}
    }

    \label{tab:benchmark_comparison}
\end{table}

General-purpose benchmarks such as AgentBench~\citep{liu2024agentbench} and GAIA~\citep{mialon2024gaia} evaluate agent capabilities across diverse environments, while MultiAgentBench~\citep{zhu2025multiagentbench} focuses more directly on multi-agent coordination and communication. Other benchmarks further study agent architectures, team sizes, and collaboration protocols~\citep{grotschla2025agentsnet, zhang2026silo, yang2026understanding}. These benchmarks primarily evaluate capabilities other than autonomous recovery of missing requirements from execution evidence.

More recent benchmarks consider incomplete specifications. SpecBench~\citep{hamblin2026specbench}  evaluates whether software agents can identify omissions, ambiguities, inconsistencies, and incorrect assumptions in software design proposals. SWE-Together~\citep{wu2026swe} and \mbox{ICAE-Bench}~\citep{peng2026icae} introduce additional requirements through user interaction during coding tasks. ProgSpec complements specification-level benchmarks by evaluating requirement satisfaction across Coding, Mathematics, and document-grounded QA, with evidence drawn from the initial context and subsequent problem solving. Table~\ref{tab:benchmark_comparison} summarizes these differences.

\section{ProgSpec: Progressively Specified Tasks Benchmark}

\subsection{Progressively Specified Tasks}
\label{Notion-PST}

We define \textbf{progressively specified tasks} as tasks where the initial user request does not explicitly state all requirements needed for successful completion. Given a request $q$, let $\mathcal{R}^{*}$ denote an idealized, fixed, evidence-grounded requirement set, and let $\mathcal{R}_{\mathrm{vis}}\subseteq\mathcal{R}^{*}$ denote those explicitly stated in the initial request. In ProgSpec, the idealized set is approximated by a human-constructed reference set. Throughout the benchmark and evaluation sections, $\mathcal{R}^{*}$ denotes this annotated reference set.

Some missing requirements may be inferred immediately, while others only become clear through reasoning, retrieved evidence, tool use, or execution feedback. Let $\hat{\mathcal{R}}_t$ denote the requirements identified by the system at repository version $t$. As execution proceeds, the task specification is progressively refined from $\hat{\mathcal{R}}_0$ to $\hat{\mathcal{R}}_1,\ldots,\hat{\mathcal{R}}_T$ as new requirements are discovered. Unlike tasks with a complete specification upfront, the system must solve the task and update its understanding of the requirements as new task specification emerges.

\subsection{Benchmark Construction}

We construct \textbf{ProgSpec} to evaluate whether a multi-agent system can satisfy requirements that are not fully stated in the initial request. Each instance contains an initial request, an annotated reference requirement set $\mathcal{R}^{*}$, and annotations describing how directly each requirement can be obtained from the initial task input.

\paragraph{Data collection.}
ProgSpec covers Coding, Mathematics, and document-grounded QA. For \textbf{Coding}, we collect repository-grounded implementation-planning tasks from public GitHub projects, using Issues, pull requests, tests, and repository history as annotation evidence for missing requirements. For \textbf{Mathematics}, we select 60 problems from MATH~\citep{hendrycksmath2021} and 30 from the text-only subset of OlympiadBench~\citep{he2024olympiadbench}. For \textbf{QA}, we construct 90 document-grounded tasks from MultiDoc2Dial~\citep{feng2021multidoc2dial}, where complete answers may require identifying additional conditions, exceptions, or constraints from the provided documents. Detailed construction procedures are provided in Appendix~\ref{appendix-progspec-construction}.

\paragraph{Specification annotation.}
We decompose each task into atomic, independently verifiable requirements and assign them to three levels based on how directly they can be obtained from the initial task input. L1 requirements are explicitly stated, L2 requirements can be inferred from the request or standard domain knowledge, and L3 requirements are typically discovered during execution by inspecting task-specific evidence, intermediate results, tool feedback, or verification. The level depends on how directly a requirement is specified, not on when the system discovers it. Appendix~\ref{appendix-annotation-guideline} provides the annotation guidelines, and Appendix~\ref{sec:annotation-quality} reports the independent human validation.

\subsection{Data Statistics}
\label{sec:data-statistics}

\begin{wraptable}{r}{0.5\linewidth}
\centering
\vspace{-0.5cm}
\caption{\textbf{Statistics of ProgSpec.} L1, L2, and L3 denote explicit, implicit, and latent requirements, respectively.}
\label{tab:progspec-stats}
\small
\resizebox{\linewidth}{!}{
\begin{tabular}{lrrrrrr}
\toprule
\textbf{Domain} & \textbf{Tasks} & \textbf{Reqs.} & \textbf{L1} & \textbf{L2} & \textbf{L3} & \textbf{Reqs./task} \\
\midrule
\textbf{Coding} & 90 & 539 & 90 & 164 & 285 & 5.99 \\
\textbf{Math} & 90 & 530 & 180 & 213 & 137 & 5.89 \\
\textbf{QA} & 90 & 482 & 90 & 97 & 295 & 5.36 \\
\midrule
\textbf{Total} & 270 & 1{,}551 & 360 & 474 & 717 & 5.74 \\
\bottomrule
\end{tabular}
}
\vspace{-0.5cm}
\end{wraptable}

ProgSpec contains 270 tasks, with 90 tasks in each domain. In the standard setting, systems receive the public task input, including a repository snapshot or document context where applicable, while requirement annotations and reference answers are withheld. As shown in Table~\ref{tab:progspec-stats}, the benchmark contains 1{,}551 atomic requirements, averaging 5.74 per task, of which 46.2\% are L3 requirements.

\paragraph{Evaluation.}
We use weighted requirement coverage as the primary metric:
\begin{equation}
\mathrm{Coverage}
=
100\times
\frac{\sum_{r_i\in\mathcal{S}} w_i}
     {\sum_{r_i\in\mathcal{R}^{*}} w_i},
\end{equation}
where $\mathcal{S}$ is the set of satisfied requirements and $w_i\in\{1,2,3\}$ denotes the weight of L1--L3 requirements. This weighting emphasizes the benchmark’s focus on satisfying implicit and latent requirements. Appendix~\ref{appendix-progspec-evaluation} gives the complete aggregation and satisfaction criteria.

\section{RepoMAS: An Issue-Driven Multi-Agent System for Progressively Specified Tasks}

\begin{figure}[!ht]
    \centering
    \includegraphics[width=\linewidth]{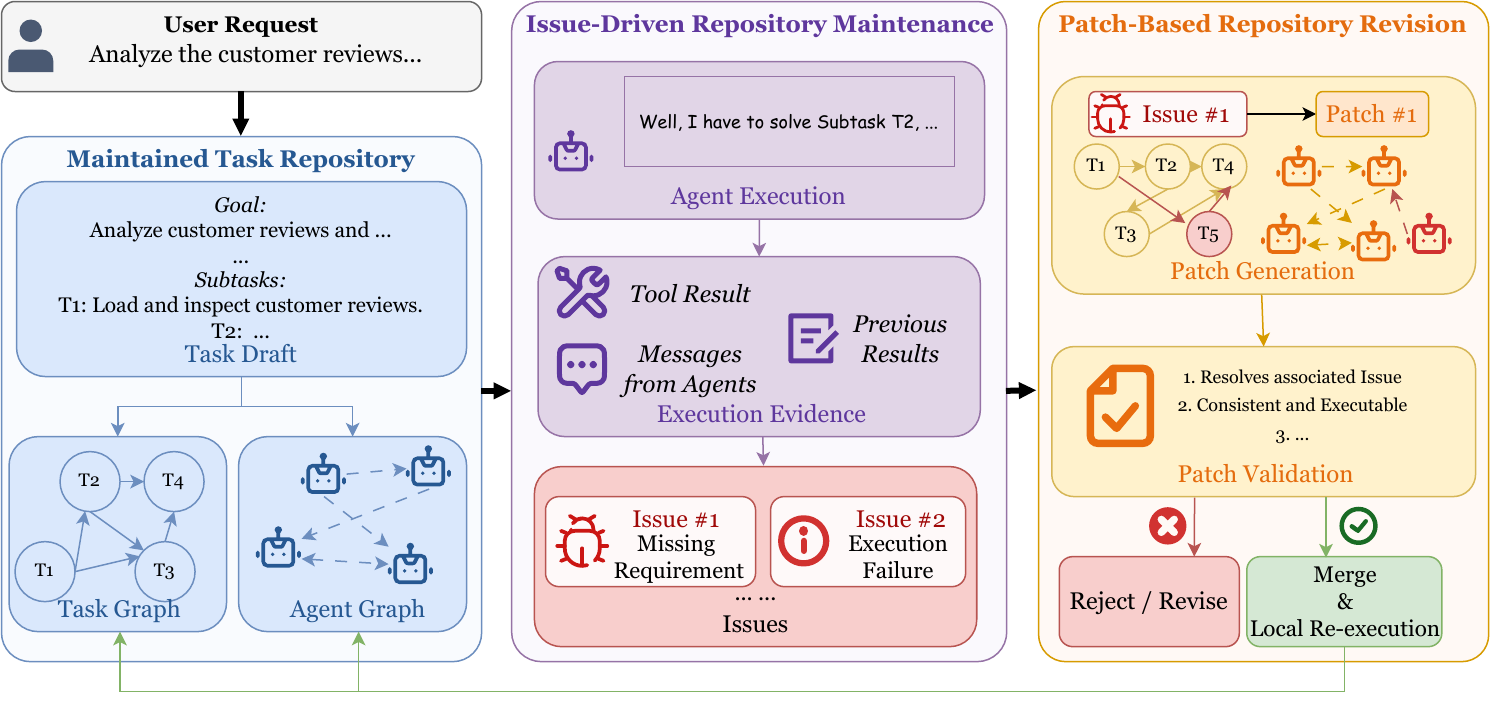}
    \caption{\textbf{An overview of RepoMAS}. RepoMAS maintains the task specification and execution structures in a shared task repository.
Execution evidence is converted into structured Issues, which drive patch-based revisions to the repository.
Candidate patches are validated before merging, and accepted revisions trigger local re-execution of only the affected tasks.}
    \label{fig-repomas}
    \vspace{-0.3cm}
\end{figure}

RepoMAS organizes multi-agent task solving around a repository-based workflow inspired by open-source development. Instead of treating the initial task plan as fixed, RepoMAS maintains a shared task state that can be inspected and revised throughout execution. Figure~\ref{fig-repomas} illustrates the overall process.

\subsection{Repository State Initialization}

At repository version $t$, RepoMAS maintains a \textbf{Task Draft}
$\mathcal{D}_t=(\hat{\mathcal{R}}_t,\mathcal{O}_t,\mathcal{V}_t)$,
which records the currently known requirements, expected outputs, and verification criteria.
The draft can be updated as new requirements are discovered.

RepoMAS then decomposes the draft into a directed acyclic \textbf{Task Graph}
$\mathcal{H}_t=(V_t^H,E_t^H)$.
Each node represents a subtask, while edges represent execution dependencies. Each subtask is also linked to the requirements it is responsible for. These links are used to locate the affected subtasks when the task specification changes.

RepoMAS also maintains an \textbf{Agent Graph}
$\mathcal{G}_t=(V_t^G,E_t^G)$,
which describes the agents in the current workflow and their communication structure. Each subtask is assigned to one or more agents through a task-agent mapping $\mathcal{M}_t$.

The repository therefore keeps both the current task structure and the current execution structure, together with the Issues discovered during execution $\mathcal{I}_t$ and the generated artifacts $\mathcal{A}_t$:
\begin{equation}
\mathcal{S}_t =
\left(
\mathcal{D}_t,
\mathcal{H}_t,
\mathcal{G}_t,
\mathcal{M}_t,
\mathcal{I}_t,
\mathcal{A}_t
\right).
\end{equation}

\subsection{Issue-Driven Repository Maintenance}

During execution, feedback may reveal problems in the current task specification or workflow. RepoMAS records each finding as a structured \textbf{Issue} before proposing a repair. Related Issues are grouped into a \textbf{repair branch}, where candidate revisions remain isolated from the accepted repository state until they are validated.

Each Issue records the problem and the evidence needed to locate and verify it, which can be modeled as
\begin{equation}
I_i=(\tau_i,\kappa_i,\phi_i,e_i),
\end{equation}
where $\tau_i$ denotes the issue type, $\kappa_i$ the affected repository component, $\phi_i$ the problem description, and $e_i$ the supporting evidence.

For a repair branch $B$, a candidate patch specifies changes to the repository components:
\begin{equation}
P_B=
\left(
\Delta\mathcal{D}_B,
\Delta\mathcal{H}_B,
\Delta\mathcal{G}_B,
\Delta\mathcal{M}_B
\right).
\end{equation}
When an Issue reveals a missing or incorrect requirement, RepoMAS revises the Task Draft and updates the affected execution structures accordingly. Issues concerning execution failures may instead require changes to the affected tasks or agent assignments without changing the Task Draft.

\subsection{Patch Validation, Merge, and Re-execution}

Before merging a candidate patch, a validator reviews it against the current repository state, the target Issues, and their supporting evidence. The review assesses whether the proposed changes address the target Issues and remain consistent with the task requirements. Rejected candidates may be revised, while accepted patches are merged into the repository. The effects of an accepted patch are subsequently checked through re-execution of the affected tasks.

Because a patch may invalidate previous results, RepoMAS re-executes only the affected part of the Task Graph. Let $C(P_B)$ denote the tasks directly affected by the patch, while artifacts from removed tasks are invalidated. Unaffected results are preserved for reuse. The re-execution set is
\begin{equation}
\mathcal{V}_{\mathrm{re}}
=
C(P_B)
\cup
\operatorname{Desc}_{\mathcal{H}_{t+1}}
\left(
C(P_B)
\right),
\end{equation}
which includes the directly affected nodes and their downstream dependent tasks in the updated graph. Outputs outside $\mathcal{V}_{\mathrm{re}}$ are reused only when their inputs, assigned requirements, and producing agents remain unchanged.

The accepted-patch path for a repair branch can be summarized as
\begin{equation}
\mathcal{S}_t
\xrightarrow{\mathrm{Execute}}
\mathcal{I}^{\mathrm{new}}_t
\xrightarrow{\mathrm{Branch/Patch}}
P_B
\xrightarrow{\mathrm{Review/Merge/Local\ Re-execution}}
\mathcal{S}_{t+1}.
\end{equation}
Here $\mathcal{I}^{\mathrm{new}}_t$ denotes newly recorded Issues, rather than the full Issue history in $\mathcal{S}_t$. A rejected patch remains outside the accepted state and may be revised. Re-execution may reveal further requirements or problems, which are again recorded as Issues. The process continues until no further revision is needed or the execution budget is reached.

\section{Experiments}

\subsection{Experimental Setup}

\textbf{Benchmarks.}
We evaluate RepoMAS on ProgSpec and 5 conventional benchmarks. Following Meta-Agent~\citep{xu2026meta}, we evaluate on the AFlow~\citep{zhang2025aflow} test splits of HumanEval~\citep{chen2021evaluating}, MBPP~\citep{austin2021program}, HotpotQA~\citep{yang2018hotpotqa}, DROP~\citep{dua-etal-2019-drop}, and GSM8K~\citep{cobbe2021training}. These benchmarks provide a broader evaluation of RepoMAS on conventional coding, reasoning, and QA tasks.

\paragraph{Baselines.}
We compare RepoMAS with GPT-4o-mini as a single-agent baseline and seven representative multi-agent systems, including  MetaGPT~\citep{hong2024metagpt}, AgentCoder~\citep{huang2023agentcoder}, G-Designer~\citep{zhang2024g}, AFlow~\citep{zhang2025aflow}, Meta-Agent~\citep{xu2026meta}, TacoMAS~\citep{xu2026tacomas}, and LATTE~\citep{mieczkowski2026improving}. For a fair comparison, all methods use GPT-4o-mini as the backbone and share the same task inputs, tool environment, and evaluation protocol, while some baseline results are taken directly from their original papers. Appendix~\ref{exp-config} provides additional details on the experimental setup and statistical significance tests.

\textbf{Metrics.}
On ProgSpec, we report weighted requirement coverage as the primary metric, which is defined in Appendix~\ref{appendix-progspec-evaluation}. On the standard benchmarks, we report pass@1 for HumanEval and MBPP, solve rate for GSM8K, and F1 for HotpotQA and DROP.

\subsection{Main Results}

    \begin{table}[!ht]
        \centering
        \vspace{-0.3cm}
\caption{\textbf{Main results with GPT-4o-mini on six benchmarks.} We report pass@1 for HumanEval and MBPP, F1 for HotpotQA and DROP, solve rate for GSM8K, and weighted requirement coverage for ProgSpec. Bold marks the best reported result in each column. * indicates results are reported in the original paper. The overall performance trend is statistically significant ($p=0.024$).}
        \resizebox{\linewidth}{!}{
        \begin{tabular}{lcccccc}
            \toprule
             \textbf{Method} & \textbf{HumanEval} &	\textbf{HotpotQA} & 	\textbf{DROP} & 	\textbf{GSM8K} & 	\textbf{MBPP} & \textbf{ProgSpec}\\
             \midrule
            GPT-4o-mini & 87.0 & 68.1 & 68.3 & 92.7 & 71.8& 38.3\\
            MetaGPT~\citep{hong2024metagpt} & 87.0 & 67.3 & 77.3 & 82.8 & 66.6 & 8.9 \\
            AgentCoder~\citep{huang2023agentcoder} & 82.4 & 54.9 & 82.0 & 88.5 & 59.2 & 12.8 \\ 
            G-Designer~\citep{zhang2024g} & 88.6 & 68.1 & 71.5 & 95.1& 71.3& 11.1\\
            AFlow~\citep{zhang2025aflow} & 94.7	&73.5&	80.6&	93.5	&83.4& 39.9\\
Meta-Agent~\citep{xu2026meta}
& 96.0$^*$ & 69.5$^*$ & 82.7$^*$ & 93.7$^*$ & 84.6$^*$ & 42.9 \\
            TacoMAS~\citep{xu2026tacomas}& 87.8 & 47.4 & 36.4 & 88.1 & 70.1 & 16.3\\
            LATTE~\citep{mieczkowski2026improving}& 90.8& 60.4&83.0&86.1&68.3&9.4 \\
            \textbf{RepoMAS (Ours)} & \textbf{99.2} &	\textbf{75.5}	& \textbf{86.3} &	\textbf{95.8} &	\textbf{87.1}& \textbf{44.7} \\
            \bottomrule
        \end{tabular}
        }
        \label{t-main-result}
    \end{table}

Table~\ref{t-main-result} reports the main results with GPT-4o-mini as the backbone. RepoMAS achieves the highest score across all six benchmarks. On ProgSpec, RepoMAS achieves 44.7 weighted requirement coverage, outperforming Meta-Agent by 1.8 points and GPT-4o-mini by 6.4 points. AFlow and Meta-Agent reach 39.9 and 42.9, respectively, while several MASs perform below the single-agent baseline. These results show that RepoMAS is more effective at handling progressively specified tasks. Appendix \ref{sec:case-study} presents three RepoMAS cases. Additional results with different backbone models are provided in Appendix~\ref{backbone-robustness}.

\section{Analysis}

\subsection{Ablation Study}

\begin{wraptable}{r}{0.6\linewidth}
\centering
\vspace{-0.5cm}
\caption{\textbf{Ablation study of RepoMAS.} Each variant removes one major component from the full framework.}
\label{tab-ablation}
\resizebox{\linewidth}{!}{
    \begin{tabular}{lccc}
    \toprule
    \textbf{Setting} & \textbf{ProgSpec} & \textbf{HumanEval} & \textbf{HotpotQA} \\
    \midrule
    \textbf{RepoMAS} & \textbf{44.7} & \textbf{99.2} & \textbf{75.5} \\
    w/o Structured Issue & 30.8 & 84.7 & 68.5 \\
    w/o Patch Validation & 33.4 & 84.0 & 59.4 \\
    w/o Structural Revision & 30.3 & 87.8 & 56.4 \\
    w/o Local Re-execution & 32.4 & 84.7 & 69.7 \\
    \bottomrule
    \end{tabular}
}
\end{wraptable}
We ablate four core components of RepoMAS using GPT-4o-mini on ProgSpec, HumanEval, and HotpotQA. \emph{w/o Structured Issue} replaces the structured Issue representation with a free-form record, while keeping the Issue stage itself. \emph{w/o Patch Validation} removes the validation step before a candidate patch is merged. \emph{w/o Structural Revision} allows the Task Draft to be updated but keeps the task and agent structures fixed. \emph{w/o Local Re-execution} disables affected-subgraph re-execution and reruns the workflow without this locality mechanism.

Table~\ref{tab-ablation} shows that removing any component reduces performance across all three benchmarks. Structural Revision has the largest effect on ProgSpec and HotpotQA. Patch Validation causes the largest drop on HumanEval, highlighting the importance of reviewing candidate revisions before they are merged. Removing structured Issues also leads to substantial degradation, showing the benefit of explicitly organizing execution feedback before repair. Local Re-execution also improves the performance, although its effect is smaller on HotpotQA.

\subsection{Effect of Structured Issue Representation}

\begin{wraptable}{r}{0.6\linewidth}
    \centering
    \small
    \vspace{-0.3cm}
    \caption{\textbf{Effect of different Issue representations.} All variants differ only in how execution feedback is represented before patch generation.}
    \label{tab:issue_representation_analysis}
    \resizebox{\linewidth}{!}{
    \begin{tabular}{lcccc}
        \toprule
        \textbf{Issue Representation}
        & \textbf{ProgSpec}
        & \textbf{HumanEval}
        & \textbf{HotpotQA}
        & \textbf{Avg.} \\
        \midrule
        Direct Revision & 30.2 & 90.8 & 68.5 & 63.2 \\
        Free-form Issue & 30.8 & 84.7 & 68.5 & 61.4 \\
        Structured Issue w/o Evidence & 27.5 & 90.8 & 69.0 & 62.5 \\
        Full Structured Issue & \textbf{44.7} & \textbf{99.2} & \textbf{75.5} & \textbf{73.1} \\
        \bottomrule
    \end{tabular}
    }
    \vspace{-0.3cm}
\end{wraptable}

We study how the representation of execution feedback affects RepoMAS using GPT-4o-mini on ProgSpec, HumanEval, and HotpotQA. \emph{Direct Revision} generates a patch directly from execution feedback without creating an Issue. \emph{Free-form Issue} first summarizes the problem in natural language before patch generation. \emph{Structured Issue w/o Evidence} uses the predefined Issue fields but omits supporting evidence. \emph{Full Structured Issue} uses the complete RepoMAS representation, where each Issue is structured and grounded in evidence. All other components remain unchanged.

Table~\ref{tab:issue_representation_analysis} shows a clear advantage for the full Issue representation across all three benchmarks. Neither an unstructured intermediate Issue nor structure alone provides the same improvement.

\subsection{Effectiveness of Patch Validation}

\begin{wraptable}{r}{0.38\linewidth}
    \centering
    \small
    \vspace{-0.8cm}
    \caption{\textbf{Effectiveness of Patch Validation on ProgSpec-Coding.}}
    \label{tab:patch-validation}
    \resizebox{\linewidth}{!}{
    \begin{tabular}{lc}
        \toprule
        \textbf{Metric} & \textbf{Value} \\
        \midrule
        Validation Precision & 62.5\% \\
        Validation Recall & 86.2\% \\
        Accepted-patch Error Rate & 37.5\% \\
        Regression Rate & 22.5\% \\
        \midrule
        Good patches: initial & 13.8\% \\
        Good patches: revised & 69.0\% \\
        \bottomrule
    \end{tabular}
    }
    \vspace{-0.5cm}
\end{wraptable}

We evaluate Patch Validation on candidate patches from ProgSpec-Coding. Each patch is labeled according to whether it resolves the target Issue without breaking the current specification, and we compare these labels with the validator's accept/revise decisions. Detailed labeling criteria and metric definitions are provided in Appendix~\ref{appendix-patch-validation}.

As shown in Table~\ref{tab:patch-validation}, Patch Validation retains most valid patches, with 86.2\% recall and 62.5\% precision. The proportion of good patches rises from 13.8\% before revision to 69.0\% after revision. This suggests that Patch Validation not only filters unreliable patches, but also helps improve patches that initially fail the review.

\subsection{Effects of Structural Revision}

\begin{wraptable}{r}{0.6\linewidth}
    \centering
    \vspace{-0.5cm}
    \caption{\textbf{Effects of Task and Agent Graph revision.}
    Each variant disables one or both forms of structural revision.}
    \label{tab:structural_revision}
    \resizebox{\linewidth}{!}{
    \begin{tabular}{lccc}
        \toprule
        \textbf{Variant} &
        \textbf{ProgSpec} &
        \textbf{HumanEval} &
        \textbf{HotpotQA} \\
        \midrule
        \textbf{Full RepoMAS} & \textbf{44.7} & \textbf{99.2} & \textbf{75.5} \\
        w/o Task Graph Revision & 25.8 & 90.1 & 55.8 \\
        w/o Agent Graph Revision & 25.8 & 88.5 & 56.9 \\
        w/o Structural Revision & 30.3 & 87.8 & 56.4 \\
        \bottomrule
    \end{tabular}
    }
    \vspace{-0.4cm}
\end{wraptable}

We examine how structural revision contributes to RepoMAS by selectively disabling updates to the Task Graph and Agent Graph. In all variants, the Task Draft can still incorporate newly discovered requirements.

Table~\ref{tab:structural_revision} shows that the full revision mechanism consistently performs best across all three benchmarks. On ProgSpec, disabling either Task Graph or Agent Graph revision reduces the score from 44.7 to 25.8. Disabling both revisions yields 30.3. This result highlights the importance of keeping the task structure and agent organization synchronized after the specification changes. Updating only one graph can leave the execution structure inconsistent, while RepoMAS jointly adapts both structures to the new specification.

\subsection{Requirement-Level Composition}
\label{sec:req-level-composition}
\begin{figure}[!ht]
    \centering
    \includegraphics[width=\linewidth]{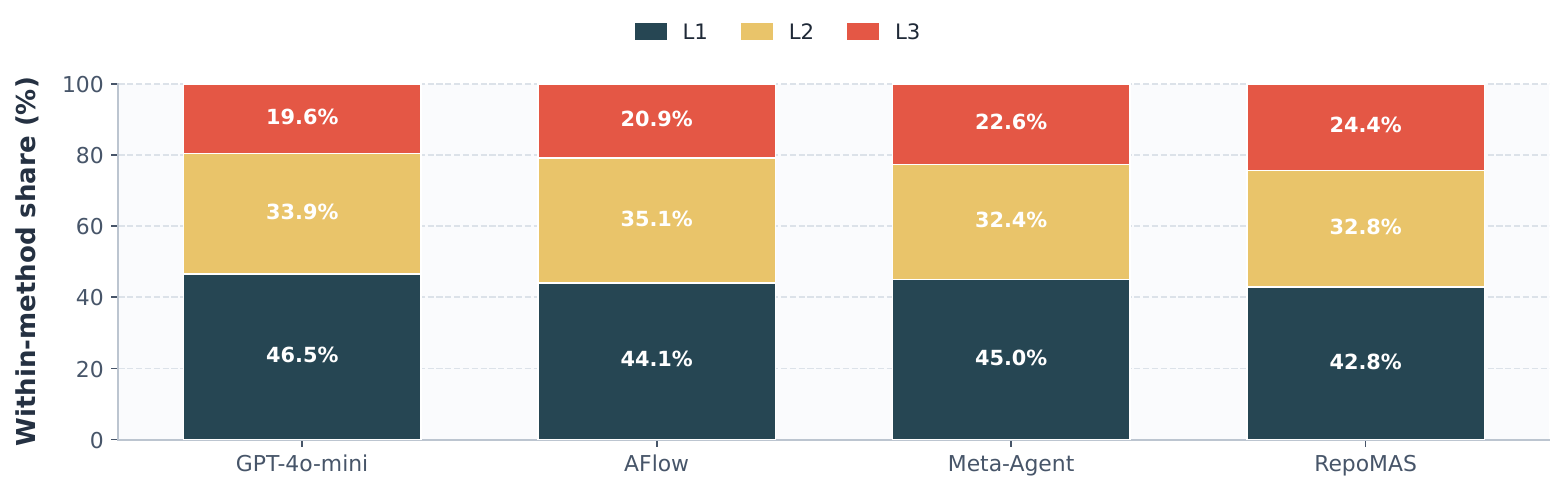}
    \caption{
\textbf{Composition of satisfied requirements by specification level on ProgSpec.} Each bar shows the within-method distribution of satisfied requirements across explicit (L1), implicit (L2), and latent (L3) requirements.
}
    \label{fig:req-level-composition}
\end{figure}

We further analyze the composition of requirements satisfied by different methods across L1, L2, and L3. Figure~\ref{fig:req-level-composition} reports the within-method distribution, normalized over the requirements satisfied by each method.

The share of L3 requirements increases from 19.6\% for the single-agent baseline to 24.4\% for RepoMAS, while the share of L1 decreases from 46.5\% to 42.8\%. L2 remains relatively stable. This suggests that RepoMAS satisfies a larger proportion of latent requirements rather than concentrating mainly on requirements explicit in the initial request.

\subsection{Requirement Discovery Continues Throughout Execution}

\begin{figure}[!ht]
    \centering

\includegraphics[width=\linewidth]{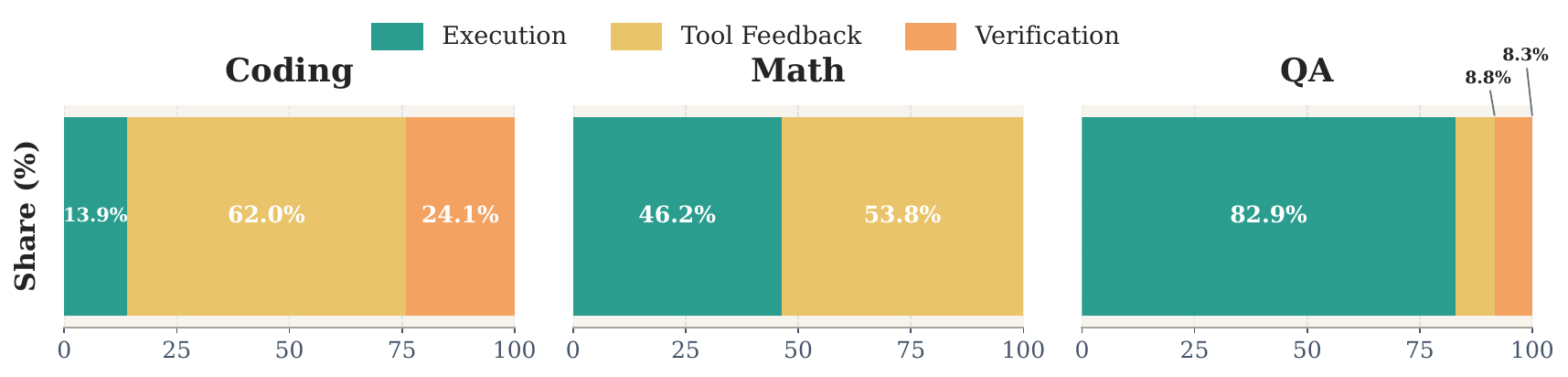}
    \caption{\textbf{First discovery stages of L3 requirements identified by RepoMAS with GPT-4o-mini on ProgSpec.}
    Percentages are normalized over discovered L3 requirements within each domain.}
    \label{fig-latent-discover-time}
    \vspace{-0.15cm}
\end{figure}

We analyze when RepoMAS first identifies L3 requirements during execution. Figure~\ref{fig-latent-discover-time} shows their distribution across \emph{Execution}, \emph{Tool Feedback}, and \emph{Final Verification}, normalized within each domain over requirements first identified in these three stages. In Mathematics, Tool Feedback is generated during execution by independently recomputing candidate results, identifying incorrect calculations or derivation steps, and requesting targeted corrections. In contrast, Final Verification checks the completed output before release and identifies requirements that remain missing or unsatisfied after execution.

The discovery pattern varies across domains. In Coding, most discoveries occur through Tool/Verifier Feedback (62.0\%), followed by Final Verification (24.1\%). In Mathematics, discoveries are split between Execution (46.2\%) and verifier feedback (53.8\%), with no requirements first identified during Final Verification. In QA, most discoveries occur during Execution (82.9\%). These results show that requirement discovery occurs at different stages
of the workflow.

\subsection{Token Cost Analysis}
\label{appendix-token-cost}

\begin{figure}[!ht]
    \centering
    \includegraphics[width=\linewidth]{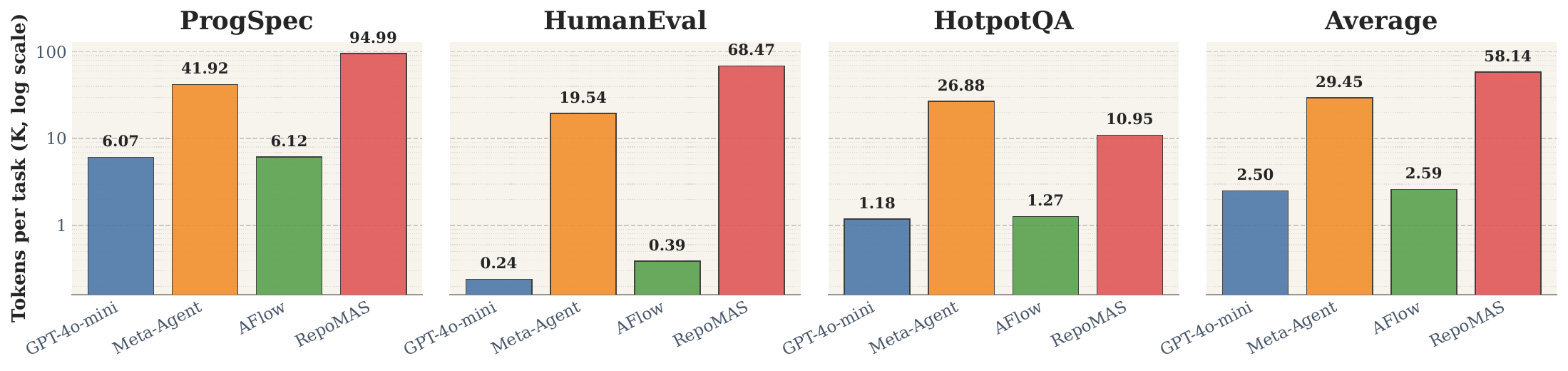}
    \vspace{-0.1cm}
    \caption{\textbf{Token consumption across three benchmarks.}
We report the average number of tokens consumed per task (K) on ProgSpec, HumanEval, and HotpotQA using a logarithmic scale.}
    \label{fig:token-cost}
    \vspace{-0.15cm}
\end{figure}

Figure \ref{fig:token-cost} reports the average token usage per task across the three benchmarks. RepoMAS consumes more tokens than the baselines due to the additional computation introduced by issue tracking, branch-based revision, patch validation, and iterative repair. The largest overhead appears on ProgSpec and HumanEval, where RepoMAS performs more extensive repository updates and validation during execution. On HotpotQA, the gap is substantially smaller. Overall, the results show that the token cost of RepoMAS is closely tied to the amount of repository maintenance required by the task. Benchmarks involving more revisions and validation steps incur higher token consumption.

\section{Conclusion}
\vspace{-0.1cm}
We study \textit{progressively specified tasks}, where task requirements may only become clear during reasoning, tool use, or execution. We introduce \textbf{ProgSpec}, a 270-task benchmark across Coding, Mathematics, and document-grounded QA, and propose \textbf{RepoMAS}, an issue-driven multi-agent framework inspired by open-source project management. RepoMAS maintains an evolving task specification through structured Issues, patches, validation, and selective re-execution. With GPT-4o-mini, RepoMAS achieves 44.7\% weighted requirement coverage on ProgSpec, 1.8 percentage points above the strongest evaluated baseline, and the highest reported scores on five conventional benchmarks. These results motivate systems that revise their explicit representation of task requirements as new evidence becomes available.

\section*{AI Use Statement}

AI tools were used for language polishing, code assistance, manuscript formatting, and limited editing of technical descriptions. They also supported routine programming and document-preparation tasks. All research decisions, methodological design, dataset construction, implementation choices, experimental analysis, and interpretation of results were carried out by the authors. All AI-assisted outputs, including code and text, were reviewed and verified by the authors before being incorporated into the final manuscript.

\bibliographystyle{iclr2027_conference}
\bibliography{document}

\appendix

\input{case_study_revised}
\section{ProgSpec Construction Details}
\label{appendix-progspec-construction}

This section describes the construction of ProgSpec.
ProgSpec contains Coding, Mathematics, and document-grounded QA tasks.
 Each gold requirement must be supported by the task context or
be discoverable during execution. Three graduate students independently annotated all 270 ProgSpec tasks. Each annotator reviewed the full set and assigned requirement-level judgments. We then compared the three annotations for every task. Cases with inconsistent labels were discussed and re-examined against the task description and supporting evidence. The final annotation was determined after resolving these disagreements, yielding a single consolidated set of labels for the benchmark.

\subsection{Coding Task Construction}

The Coding split contains repository-grounded implementation-planning tasks
built from public Python projects on GitHub. Given a change request and a fixed
repository snapshot, the system must produce an implementation plan that
captures the behavior required by the task and identifies the relevant parts of
the codebase. We evaluate the plan against atomic requirements rather than by
executing a code patch.

We select repositories that contain multiple interacting files or modules and require reasoning across component boundaries. Each repository must also provide development records, such as issues, pull requests, commits, or tests, that can be used to reconstruct task requirements. We exclude repositories whose execution depends on unavailable external services, complex environment configurations, or insufficient development evidence. For each repository, we restrict the task scope to the files and modules directly involved in the target change, while retaining their relevant dependency relations.

The 12 repositories used in ProgSpec are:

\begin{quote}
\textit{attrs}, \textit{cattrs}, \textit{dateparser}, \textit{filelock},
\textit{pendulum}, \textit{pluggy}, \textit{pydantic-settings},
\textit{python-dotenv}, \textit{requests-cache}, \textit{tenacity},
\textit{tomlkit}, and \textit{typer}.
\end{quote}

Each task is created from a natural-language change request and a fixed
repository snapshot. For requirements that are not explicit in the request,
we record the evidence used during annotation and separately track whether that
evidence is available to the solving system. Annotation may draw on repository
history, while inference-time evidence comes from the released codebase,
available tests, and observable execution feedback.

We retain only requirements that are necessary for the requested change and
recoverable from the allowed task context. Historical preferences or optional
design choices are excluded when the current repository does not provide enough
evidence to make them mandatory.

\subsection{Mathematics Task Construction}

The Mathematics split contains 60 problems from MATH~\citep{hendrycksmath2021} and 30 problems from the text-only subset of OlympiadBench~\citep{he2024olympiadbench}.

We select problems with well-defined final answers and enough internal structure to support multiple independently verifiable requirements. 
Problems that depend on images, external tools, or ambiguous assumptions are excluded.

For each problem, we inspect both the problem statement and its reference solution. We derive candidate requirements that are necessary for a complete and correct answer without treating the reference derivation as the only valid solution path.
The resulting requirements may include the requested target, domain restrictions, assumptions, intermediate conditions, boundary cases, or checks needed to ensure that the final result is valid.

We avoid treating individual derivation steps as separate requirements. Direct implications of the statement and standard mathematical domain conditions are labeled L2. L3 requirements concern task-specific conditions whose necessity is established through further analysis or verification, rather than through the direct application of a standard domain rule. Such conditions must constrain the validity or completeness of the solution independently of the chosen derivation. Requirements tied only to one reference solution path are excluded, and alternative correct derivations are accepted.

\subsection{QA Task Construction}

The QA split contains 90 document-grounded tasks constructed from
MultiDoc2Dial~\citep{feng2021multidoc2dial}. Each task provides an initial
user request together with the documents needed to answer it.

We select tasks where a complete answer requires combining information from
multiple parts of the documents rather than retrieving a single sentence.
Some requirements are not stated directly in the user request, but can be
recovered from the provided document context.

Only the initial user request and document context are given to the system.
Later dialogue turns are not used as task input or as sources of gold
requirements. This ensures that latent requirements come from the available
documents rather than from later user clarification.

For each gold requirement, we record the supporting document passage.
Requirements without support in the provided documents are excluded.

\section{ProgSpec Annotation Guidelines}
\label{appendix-annotation-guideline}

\subsection{Atomic Requirement Annotation}

For each task, annotate a set of atomic requirements that together define what
a successful solution must satisfy. These requirements form the reference set
$\mathcal{R}^{*}$.

Each requirement must satisfy the following rules:

\begin{itemize}
    \item \textbf{Atomicity:} Write one requirement per item. Split independent
    conditions into separate requirements.
    
    \item \textbf{Verifiability:} Phrase the requirement so that its satisfaction
    can be checked from the final output or execution result.
    
    \item \textbf{Grounding:} Include a requirement only when it is supported by
    the task statement or the evidence available for that task.
\end{itemize}

Annotate requirements that are necessary for successful task completion.
These may cover expected behavior, boundary conditions, reasoning constraints,
or facts that must appear in the final answer.

Do not annotate optional improvements, stylistic preferences, or implementation
choices that are not required by the task.

\subsection{Requirement Levels}

Assign each requirement to one of the following levels according to its supporting evidence and derivation:
\begin{enumerate}
    \item \textbf{L1 — Explicit}: The requirement is explicitly stated in the initial user request.
    \item \textbf{L2 — Implicit}: The requirement is a direct consequence of the request or a standard domain condition, such as the domain restriction of a stated mathematical expression.
    \item \textbf{L3 — Latent}: Establishing the requirement requires locating and connecting task-specific evidence, or carrying out task-specific analysis or verification. Relevant evidence may come from supplied repository files, document passages, tool outputs, or intermediate reasoning results.
\end{enumerate}

Classify directly stated requirements as L1 and direct implications or standard domain conditions as L2, even if a system discovers them later. For each L3 requirement, record the specific evidence or derivation that establishes it.

\subsection{Annotation Procedure}

Each task is annotated using the following procedure:

\begin{enumerate}
    \item Inspect the initial task request and the available supporting evidence.
    \item Identify the conditions that must hold for the task to be considered successfully completed.
    \item Decompose these conditions into atomic and independently verifiable requirements.
    \item Assign each requirement an L1, L2, or L3 label according to the evidence criteria in the requirement-level definitions.
    \item Record both annotation evidence and inference-accessible evidence, together with the action needed to acquire the latter.
    \item Review the final requirement set for missing, duplicated, or overly broad requirements.
\end{enumerate}

\subsection{Annotation Examples}

Table~\ref{tab:annotation-examples} shows representative examples of the three
requirement levels.

\begin{table}[!ht]
\centering
\caption{\textbf{Examples of ProgSpec requirement levels.}}
\label{tab:annotation-examples}
\small
\begin{tabular}{p{0.2\linewidth}p{0.8\linewidth}}
\toprule
\textbf{Level} & \textbf{Example and evidence} \\
\midrule
L1 (Explicit) &
The user asks the function to return a JSON object. This requirement is stated
directly in the request. \\

L2 (Implicit) &
The task asks for the real-valued expression $\log(x-1)$ to be evaluated.
The condition $x>1$ follows directly from the domain of the real logarithm, although it
is not stated in the request. \\

L3 (Latent) &
The request does not specify how empty input should be handled. Inspection of
an available repository test shows that empty input must return an empty result
rather than raise an exception. \\
\bottomrule
\end{tabular}
\end{table}

\subsection{Quality Control}

After the initial annotation, each task should be reviewed by another annotator.
The reviewer checks the following aspects:

\begin{itemize}
    \item \textbf{Validity:} Every requirement must be necessary for successful
    task completion and supported by available evidence.

    \item \textbf{Atomicity:} Each item should express one independently
    verifiable requirement without unnecessary overlap with other items.

    \item \textbf{Evidence:} The recorded evidence must be sufficient to support
    the annotated requirement.

    \item \textbf{Level:} The L1--L3 label must follow the requirement-level
    definitions in the annotation guidelines.
\end{itemize}

Requirements that fail any of these checks are revised or removed before the
final annotation is accepted.

\section{ProgSpec Evaluation Details}
\label{appendix-progspec-evaluation}

\subsection{Weighted Requirement Coverage}

ProgSpec evaluates the final artifact against the atomic requirements of each
task. Requirements are weighted by their annotation level:

\begin{equation}
w_i =
\begin{cases}
1, & r_i \in \mathcal{R}_{\mathrm{L1}},\\
2, & r_i \in \mathcal{R}_{\mathrm{L2}},\\
3, & r_i \in \mathcal{R}_{\mathrm{L3}}.
\end{cases}
\end{equation}

Let $\mathcal{S}_q \subseteq \mathcal{R}^{*}_q$ denote the requirements
satisfied by the final artifact for task $q$. Its weighted requirement coverage
is

\begin{equation}
\mathrm{Coverage}(q)
=
100\times
\frac{
\sum_{r_i\in\mathcal{S}_q} w_i
}{
\sum_{r_i\in\mathcal{R}^{*}_q} w_i
}.
\end{equation}

For a collection of tasks $\mathcal{Q}$, we pool requirement weights across
tasks:

\begin{equation}
\mathrm{Coverage}_{\mathcal{Q}}
=
100\times
\frac{
\sum_{q\in\mathcal{Q}}\sum_{r_i\in\mathcal{S}_q}w_i
}{
\sum_{q\in\mathcal{Q}}\sum_{r_i\in\mathcal{R}^{*}_q}w_i
}.
\end{equation}

The weighting gives greater emphasis to requirements that are not explicitly
stated in the initial request. Reported domain and overall scores are computed
from the pooled requirement weights rather than by averaging task scores. 

\subsection{Requirement Satisfaction}

Each atomic requirement is evaluated against the system's final artifact.
The evaluator uses the requirement and its supporting evidence to determine
whether the required condition is satisfied.

For Coding, evaluation is performed on the final implementation plan.
For Mathematics, equivalent correct derivations and answers are accepted.
For QA, the final answer must contain the information supported by the supplied
documents. A requirement is marked as satisfied only when the final artifact
meets it without contradicting the available evidence.

\section{Patch Validation Evaluation}
\label{appendix-patch-validation}

We evaluate Patch Validation on candidate patches collected from
ProgSpec-Coding. The online validator decides whether each candidate patch is
accepted or returned for revision. We compare this decision with an offline
quality label.

\paragraph{Patch quality.}
A patch is labeled \emph{good} when it resolves the target Issue and produces
a revision consistent with the task requirements and available evidence.
Previously established requirements must remain satisfied unless new evidence
shows that they should be revised. All other patches are labeled \emph{bad}.
The offline evaluator has access to the annotated requirements, while the
online validator does not.

\paragraph{Validation metrics.}
Let $N_{\mathrm{AG}}$ and $N_{\mathrm{AB}}$ denote accepted good and accepted
bad patches, and let $N_{\mathrm{RG}}$ denote rejected good patches. We compute

\begin{equation}
\mathrm{Precision}
=
\frac{N_{\mathrm{AG}}}
     {N_{\mathrm{AG}}+N_{\mathrm{AB}}},
\qquad
\mathrm{Recall}
=
\frac{N_{\mathrm{AG}}}
     {N_{\mathrm{AG}}+N_{\mathrm{RG}}}.
\end{equation}

The Accepted-patch Error Rate is the fraction of accepted patches labeled bad,
and is therefore $1-\mathrm{Precision}$.

We also measure whether an accepted patch breaks requirements that were valid
before revision:

\begin{equation}
\mathrm{Regression\ Rate}
=
\frac{N_{\mathrm{AR}}}{N_{\mathrm{A}}},
\end{equation}

where $N_{\mathrm{AR}}$ is the number of accepted patches that introduce a
regression and $N_{\mathrm{A}}$ is the total number of accepted patches.
All metrics are reported as percentages.

\section{Backbone Robustness}
\label{backbone-robustness}

To examine whether the performance of RepoMAS depends on a specific backbone model, we evaluate three additional backbones, which are GPT-5.5, Gemini 3.7 Flash, and DeepSeek-V4-Flash. Table~\ref{tab:backbone_robustness} reports the results.

Within each backbone group, we use the same benchmark splits and evaluation metrics as in the main experiment. Except for the backbone model, the evaluated configurations use the same prompts and benchmark settings.

Results across additional backbones further demonstrate the applicability of RepoMAS beyond GPT-4o-mini, with improvements observed across most of the model families and tasks.

\begin{table*}[!ht]
    \centering
        \caption{
    \textbf{Backbone robustness results on six benchmarks.}
    We repeat the main experiments using different backbone models while keeping the methods and evaluation settings unchanged.
    All methods within each group use the same backbone model.
    Bold marks the highest score in each column.
    }
    \scriptsize
    \resizebox{\textwidth}{!}{
    \begin{tabular}{llcccccc}
        \toprule
        \textbf{Backbone} &
        \textbf{Method} &
        \textbf{HumanEval} &
        \textbf{HotpotQA} &
        \textbf{DROP} &
        \textbf{GSM8K} &
        \textbf{MBPP} &
        \textbf{ProgSpec} \\
        \midrule

  \multirow{9}{*}{GPT-5.5}
  & GPT-5.5 & 99.24 & 59.38 & 39.12 & 93.46 & 83.28 & 92.08 \\
  & MetaGPT~\citep{hong2024metagpt} & 98.47 & 32.88 & 3.38 & 71.18 & 81.23 & 89.95 \\
  & AgentCoder~\citep{huang2023agentcoder} & 99.24 & 50.62 & 33.75 & 92.61 & 83.58 & 93.02 \\
  & G-Designer~\citep{zhang2024g} & \textbf{100.00} & 57.00 & 36.50 & 93.36 & 84.46 & 94.20 \\
  & AFlow~\citep{zhang2025aflow} & 99.24 & 57.75 & 39.12 & \textbf{94.22} & 83.58 & 93.34 \\
  & Meta-Agent~\citep{xu2026meta} & 98.47 & 54.37 & 25.50 & 92.89 & 80.06 & 89.50 \\
  & TacoMAS~\citep{xu2026tacomas} & 99.24 & 56.62 & 34.38 & 92.70 & 81.23 & 92.55 \\
  & LATTE~\citep{mieczkowski2026improving} & 99.24 & 46.00 & 28.62 & 92.32 & 82.11 & 92.06 \\
  & \textbf{RepoMAS} & \textbf{100.00} & \textbf{70.00} & \textbf{66.89} & 82.84 & \textbf{94.13} & \textbf{94.45} \\
\midrule
  \multirow{9}{*}{Gemini 3.7 Flash}
  & Gemini 3.7 Flash & \textbf{100.00} & 65.00 & 49.50 & 95.45 & 89.44 & 67.48 \\
  & MetaGPT~\citep{hong2024metagpt} & 97.71 & 44.38 & 25.25 & 81.04 & 71.26 & 92.08 \\
  & AgentCoder~\citep{huang2023agentcoder} & \textbf{100.00} & 63.62 & 45.00 & 61.04 & \textbf{90.91} & 90.85 \\
  & G-Designer~\citep{zhang2024g} & \textbf{100.00} & 65.88 & 47.38 & 78.01 & 89.44 & 92.71 \\
  & AFlow~\citep{zhang2025aflow} & \textbf{100.00} & 65.50 & 48.75 & 94.50 & 88.56 & 89.95 \\
  & Meta-Agent~\citep{xu2026meta} & \textbf{100.00} & 63.50 & 40.62 & 93.55 & 87.10 & 90.14 \\
  & TacoMAS~\citep{xu2026tacomas} & \textbf{100.00} & 65.50 & 47.75 & 95.64 & 89.15 & 89.41 \\
  & LATTE~\citep{mieczkowski2026improving} & \textbf{100.00} & 61.50 & 47.38 & 67.11 & 89.15 & \textbf{94.46} \\
  & \textbf{RepoMAS} & \textbf{100.00} & \textbf{72.63} & \textbf{67.25} & \textbf{97.44} & 90.62 & 94.13 \\
        \midrule
  \multirow{9}{*}{DeepSeek-V4-Flash}
  & DeepSeek-V4-Flash & 99.24 & \textbf{58.13} & 40.62 & 93.27 & 82.40 & 70.63 \\
  & MetaGPT~\citep{hong2024metagpt} & 99.24 & 38.38 & 17.38 & 80.09 & 80.94 & 66.37 \\
  & AgentCoder~\citep{huang2023agentcoder} & 97.71 & 55.38 & 36.62 & 92.99 & 82.70 & 67.45 \\
  & G-Designer~\citep{zhang2024g} & 96.95 & 54.62 & 37.12 & 91.75 & 81.82 & 67.86 \\
  & AFlow~\citep{zhang2025aflow} & 99.24 & 55.88 & 41.50 & 92.61 & 80.94 & 63.93 \\
  & Meta-Agent~\citep{xu2026meta} & \textbf{100.00} & 49.88 & 28.62 & 89.95 & 81.23 & 68.59 \\
  & TacoMAS~\citep{xu2026tacomas} & 99.24 & 54.00 & 35.12 & 91.28 & \textbf{82.99} & 65.80 \\
  & LATTE~\citep{mieczkowski2026improving} & 99.24 & 46.12 & 22.75 & 83.70 & 81.52 & 67.72 \\
  & \textbf{RepoMAS} & \textbf{100.00} & 57.75 & \textbf{42.37} & \textbf{94.98} & 80.94 & \textbf{73.72} \\
        \bottomrule
    \end{tabular}
    }

    \label{tab:backbone_robustness}
\end{table*}

\section{Annotation Quality and Human Validation}
\label{sec:annotation-quality}

We conduct an independent human review of a fixed ProgSpec annotation snapshot.
The review covers 90 randomly sampled tasks and 500 requirements. Five graduate students who were not involved in the original annotation conduct the review.

\paragraph{Review protocol.}
At the requirement level, reviewers first determine whether each requirement is
necessary for completing the task. They then check whether it is independently
testable, supported by the available evidence, and stated unambiguously.
These judgments are used to measure requirement validity, independence,
evidence sufficiency, and ambiguity.

At the task level, reviewers examine the annotation as a whole. They check
whether the requirement set is supported by the task context, sufficiently
complete, properly decomposed into atomic requirements, and assigned to the
appropriate L1--L3 levels. We compare these judgments with the original
annotations to measure annotation--review agreement.

\paragraph{Results.}
Table~\ref{tab:annotation-quality} summarizes the human validation results.
Among the 500 reviewed requirements, 477 (95.4\%) are judged valid,
491 (98.2\%) satisfy the independence criterion, and 487 (97.4\%) have
sufficient supporting evidence. Only one requirement is judged ambiguous.

Across the four task-level judgments for 90 tasks, the original annotations
agree with the independent review in 329 of 360 cases, corresponding to
91.4\% agreement and a Gwet's AC1 of 0.91. Agreement is highest for contextual
support at 98.9\%, followed by completeness and atomicity at 90.0\%, and
requirement-level assignment at 86.7\%.

Overall, the review shows that ProgSpec requirements are consistently grounded
in task evidence and can be evaluated as independent task conditions.

\begin{table}[!ht]
    \centering
    \caption{
    \textbf{Human validation of ProgSpec annotations.}
    Results are computed over 90 tasks and 500 annotated requirements.
    }
    \label{tab:annotation-quality}
    \resizebox{\linewidth}{!}{
    \begin{tabular}{lccc}
        \toprule
        \textbf{Metric} &
        \textbf{Score} &
        \textbf{95\% CI} &
        \textbf{Evaluation Unit} \\
        \midrule
        Annotation--review agreement
            & 91.4\% ($\mathrm{AC1}=0.91$)
            & [88.0, 93.9]
            & 329 / 360 judgments \\

        Requirement validity
            & 95.4\%
            & [93.2, 96.9]
            & 477 / 500 requirements \\

        Requirement independence
            & 98.2\%
            & [96.6, 99.1]
            & 491 / 500 requirements \\

        Evidence sufficiency
            & 97.4\%
            & [95.6, 98.5]
            & 487 / 500 requirements \\

        Ambiguity rate
            & 0.2\%
            & [0.0, 1.1]
            & 1 / 500 requirements \\
        \bottomrule
    \end{tabular}
    }
\end{table}

\section{Experiment Configuration}
\label{exp-config}

This section describes the common experimental settings used for RepoMAS.
Unless otherwise specified, all experiments use the full RepoMAS framework,
including structured Issues, patch validation, structural revision, and local
re-execution.

\subsection{Execution Environment}

Experiments are run on Linux with Python~3.11. Model inference is served
through an OpenAI-compatible API, while the RepoMAS runtime itself does not
require GPU computation. Structured model outputs are constrained by JSON
schemas to ensure that intermediate artifacts can be parsed and reused by
subsequent stages. All benchmark inputs are fixed before inference. ProgSpec-Coding additionally
provides a repository snapshot for each task.

\subsection{Runtime Settings}

Unless otherwise specified, RepoMAS uses at most three execution--revision
iterations per task and one task attempt. Model outputs are limited to 4,096
tokens per model call. ProgSpec runs use temperature 0, while other
benchmark runs use temperature 0.2. The evaluation settings for each benchmark are summarized in Table~\ref{tab:eval-settings}.

\begin{table}[!ht]
\centering
\small
\caption{\textbf{Evaluation settings across benchmarks.} We report the temperature, maximum number of generated tokens, and maximum number of iterations used for each benchmark.}
\label{tab:eval-settings}
\begin{tabular}{lccc}
\toprule
\textbf{Benchmark} &
\textbf{Temperature} &
\textbf{Max Tokens} &
\textbf{Max Iterations} \\
\midrule
HumanEval & 0.2 & 4096 & 3 \\
HotpotQA / DROP / GSM8K / MBPP & 0.2 & 4096 & 3 \\
ProgSpec--Coding & 0.0 & 4096 & 3 \\
ProgSpec--Math / QA & 0.0 & 4096 & 3 \\
\bottomrule
\end{tabular}
\end{table}

For the additional-backbone experiments, we keep the RepoMAS configuration
fixed wherever possible and change only the backbone model. These runs use the
same full Issue representation, patch validation, structural revision, and
local re-execution procedure.

\subsection{Prompt Protocol}

All model calls use structured prompts with explicit output schemas. Each call
receives the original task together with the information required by its
current Task Graph node. Depending on the stage, this may include upstream
artifacts, open Issues, or the current candidate under revision. Gold
requirements, hidden tests, and reference answers are never included in
generation prompts.

The initialization stage asks the model to produce a structured Task Draft from
the public request. RepoMAS then compiles this draft into the Task Graph and
Agent Graph. Worker agents execute individual subtasks and return structured
artifacts. When execution reveals a problem, the corresponding evidence is
recorded as an Issue and used to construct a candidate patch.

The final synthesizer receives the accepted artifacts produced during
execution and generates the final response to the original task. Patch
Validation is performed independently of the contributor that proposed the
revision. An accepted patch updates the repository state, after which only the
affected part of the workflow is re-executed.

\subsection{Evaluator Protocol}
After inference, ProgSpec is evaluated by a separate post-hoc LLM judge using \texttt{GPT-5.5}. For each task, the judge receives the public task input, evaluator-only atomic requirements, the reference answer when available, and the system response. It returns a structured verdict for each requirement, including a binary coverage decision and supporting evidence from the response.

\subsection{Statistical Significance}
\label{appendix:significance}

To systematically assess the statistical significance of RepoMAS's improvements, we conduct hypothesis tests across all datasets and all baseline methods. Under the same evaluation setting, each experiment is repeated 10 times, with RepoMAS and the corresponding baseline evaluated on the same task set in each run, yielding paired performance measurements. For each dataset--baseline pair, we apply a two-sided paired $t$-test to the 10 paired results.

For each comparison, the null hypothesis assumes no difference in mean performance between RepoMAS and the corresponding baseline. We use a significance level of $\alpha=0.05$ and report the resulting $p$-values across all dataset--baseline combinations. This analysis examines whether the observed performance gains of RepoMAS are consistently supported by statistical evidence rather than being attributable to run-to-run variation.

\end{document}

%% file: math_commands.tex
\usepackage{amsmath,amsfonts,bm}

\def\eqref#1{equation~\ref{#1}}

\def\1{\bm{1}}

\DeclareMathAlphabet{\mathsfit}{\encodingdefault}{\sfdefault}{m}{sl}
\SetMathAlphabet{\mathsfit}{bold}{\encodingdefault}{\sfdefault}{bx}{n}



%% file: case_study_revised.tex

\definecolor{humanblue}{RGB}{0,102,139}
\definecolor{issuered}{RGB}{186,46,52}
\definecolor{patchgreen}{RGB}{26,127,55}
\definecolor{reviewpurple}{RGB}{110,80,180}
\definecolor{graphblue}{RGB}{9,105,218}
\definecolor{cardbg}{RGB}{250,250,248}
\definecolor{framegray}{RGB}{210,210,210}

\newcommand{\rmascasecard}[3]{%
  \par\noindent
  \begin{tikzpicture}
    \node[
      draw=#1!75!black,
      fill=cardbg,
      line width=0.55pt,
      rounded corners=2pt,
      inner sep=4pt,
      text width=\dimexpr\linewidth-10pt\relax,
      align=left,
      anchor=north
    ] {\scriptsize
      {\color{#1}\bfseries #2}\par\smallskip
      #3
    };
  \end{tikzpicture}\par\vspace{2pt}%
}

\section{Case Study}
\label{sec:case-study}

Figure~\ref{fig:repomas-casestudy} shows three ProgSpec-Coding examples in
which execution reveals requirements that are not stated in the initial
request. In the first two cases, RepoMAS records the missing requirement as an
Issue, revises the affected result, and locally re-executes the relevant part
of the workflow after validation. The third case illustrates requirement
discovery from repository structure, while the later accepted repair addresses
a separate documentation and testing gap.

\begin{figure*}[t]
\centering
\setlength{\fboxrule}{0.6pt}
\setlength{\fboxsep}{6pt}
\fcolorbox{framegray}{white}{%
\begin{minipage}{\dimexpr\linewidth-2\fboxrule-2\fboxsep\relax}
\raggedright
\scriptsize

\begin{minipage}[t]{0.315\linewidth}
\vspace{0pt}
{\color{humanblue}\bfseries Stale soft locks}\par
\smallskip
\textbf{Request.} Add configurable expiration for stale soft lock files.
\par\medskip

\rmascasecard{graphblue}{Missing specification}{%
The request does not define the parameter name, type, unit, scope, or default.}

\rmascasecard{issuered}{Evidence / Issue}{%
Repository inspection reveals a concrete contract for the option and shows
that the current answer does not state it.}

\rmascasecard{patchgreen}{Revision}{%
The revised answer specifies
\texttt{stale\_expiration: Optional[float]}, measured in seconds,
stored per lock, with default \texttt{None}.}

\rmascasecard{reviewpurple}{Validation}{%
The first candidate is returned for revision.
The revised candidate is approved and merged.}

\rmascasecard{graphblue}{Local re-execution}{%
Only the affected subtasks and validator are rerun.
Post-merge verification preserves the revised contract.}
\end{minipage}%
\hfill
\begin{minipage}[t]{0.315\linewidth}
\vspace{0pt}
{\color{humanblue}\bfseries Incomplete calendar dates}\par
\smallskip
\textbf{Request.} Reject strings that do not contain a complete calendar date.
\par\medskip

\rmascasecard{graphblue}{Missing specification}{%
The request does not define how timestamps, two-digit years,
ISO week dates, or relative expressions should be treated.}

\rmascasecard{issuered}{Evidence / Issue}{%
Execution exposes ambiguous cases.
Inferred date fields are not directly observable, while relative expressions
such as ``yesterday'' still produce valid dates.}

\rmascasecard{patchgreen}{Revision}{%
The revised result clarifies that timestamps and two-digit years count,
ISO week dates require a weekday, and relative expressions remain accepted.}

\rmascasecard{reviewpurple}{Validation}{%
The incomplete candidate is returned for revision.
The clarified version is then approved and merged.}

\rmascasecard{graphblue}{Local re-execution}{%
Only the affected date-handling subtasks and validator are rerun.}
\end{minipage}%
\hfill
\begin{minipage}[t]{0.315\linewidth}
\vspace{0pt}
{\color{humanblue}\bfseries ISO interval forms}\par
\smallskip
\textbf{Request.} Parse ISO~8601 interval expressions into Pendulum intervals.
\par\medskip

\rmascasecard{graphblue}{Missing specification}{%
The request does not say whether repeating forms such as
\texttt{R[n]/...} are in scope.}

\rmascasecard{issuered}{Evidence / Issue}{%
Repository inspection shows that \texttt{Interval} stores only a start and
an end, which cannot directly represent recurrence.}

\rmascasecard{patchgreen}{Discovered requirement}{%
The supported scope is narrowed to start/end, start/duration, and
duration/end unless a separate recurrence representation is introduced.}

\rmascasecard{reviewpurple}{Validation}{%
This requirement remains in the accepted result.
A later validated patch addresses a separate documentation and testing gap.}

\rmascasecard{graphblue}{Local re-execution}{%
The later repair reruns only the affected documentation and testing tasks;
the interval-form requirement remains unchanged.}
\end{minipage}

\end{minipage}%
}

\caption{\textbf{Examples of requirement discovery in ProgSpec-Coding.}
Execution and repository evidence expose requirements that are absent from the
initial requests. In the first two cases, the discovered requirements trigger
revision, validation, and local re-execution. In the Pendulum case, the
interval-scope requirement is discovered and retained, while the later
validated repair addresses a separate gap.}
\label{fig:repomas-casestudy}
\end{figure*}

The cases illustrate that specification discovery can arise from different
forms of execution evidence. RepoMAS makes these missing requirements explicit
and incorporates them into subsequent reasoning without restarting the full
workflow.